\documentclass{article}

\usepackage{microtype}
\usepackage{graphicx}
\usepackage{subfigure}
\usepackage{booktabs} 
\usepackage{eurosym}

\usepackage{hyperref}
\usepackage{xcolor}

\definecolor{caribbeangreen}{rgb}{0.0, 0.8, 0.6}

\usepackage[accepted]{icml2020}

\icmltitlerunning{Avoiding Extra-legal Factors in Decision made by Judges and
Not Understandable AI Models}

\begin{document}

\twocolumn[ \icmltitle{Predicting Court Decisions for Alimony: Avoiding
Extra-legal Factors \\ in Decision made by Judges and Not Understandable AI
Models}



\icmlsetsymbol{equal}{*}

\begin{icmlauthorlist}

\icmlauthor{Fabrice Muhlenbach}{udl,labhc} \icmlauthor{Long Nguyen
Phuoc}{lyon2,msh} \icmlauthor{Isabelle Sayn}{msh,cmw}

\end{icmlauthorlist}

\icmlaffiliation{udl}{Universit\'{e} de Lyon}
\icmlaffiliation{labhc}{UJM-Saint-Etienne, CNRS, Laboratoire Hubert Curien, UMR
5516, 18 rue du Professeur Beno\^{\i}t Lauras, F-42023 Saint Etienne}
\icmlaffiliation{cmw}{CNRS, Universit\'{e} de Lyon, Centre Max Weber, UMR 5283}
\icmlaffiliation{msh}{Maison des Sciences de l'Homme, Lyon St-\'{E}tienne, 14
avenue Berthelot, F-69363 Lyon cedex 07} \icmlaffiliation{lyon2}{Universit\'{e}
Lumi\`{e}re Lyon 2} \icmlcorrespondingauthor{Fabrice
Muhlenbach}{fabrice.muhlenbach@univ-st-etienne.fr}

\icmlkeywords{Machine Learning, ICML}

\vskip 0.3in
]



\printAffiliationsAndNotice{\icmlEqualContribution} 

\begin{abstract}
The advent of machine learning techniques has made it possible to obtain
predictive systems that have overturned traditional legal practices. However,
rather than leading to systems seeking to replace humans, the search for the
determinants in a court decision makes it possible to give a better
understanding of the decision mechanisms carried out by the judge. By using a
large amount of court decisions in matters of divorce produced by French
jurisdictions and by looking at the variables that allow to allocate an alimony
or not, and to define its amount, we seek to identify if there may be
extra-legal factors in the decisions taken by the judges. From this
perspective, we present an explainable AI model designed in this purpose by
combining a classification with random forest and a regression model, as a
complementary tool to existing decision-making scales or guidelines created by
practitioners.
%
\end{abstract}

\section{Introduction}

Machine learning --the study of algorithms that allow computer programs to
automatically improve through experience \cite{Mitchell__1997}-- has brought
artificial intelligence to the forefront in the past decade, in particular
thanks to new techniques such as deep learning
\cite{Sejnowski__The_Deep_Learning_Revolution__2018}. Since then,
classification or clustering techniques have really improved. Effective
AI-based applications --or ``classifiers''-- have been able to be realized in
very diverse fields, whether for pattern recognition or decision support
systems, and they are able to perform complex tasks in place of humans.

Machine learning algorithms are used in finance, medicine, and criminal
justice, and therefore they can have a deep impact on society. With the recent
success of AI applications in the private and public domain, legal
professionals are now interested in artificial intelligence, especially since
many startups disrupt the legal market space by seeking to benefit of these new
AI techniques \cite{Bex_et_al_AI4J__2017}.

However, the arrival of these new techniques has brought a number of ethical
issues. Firstly, machine learning and data mining techniques are capable of
exploiting personal and legal data that are more and more easily accessible on
the Internet, leading to questions about privacy preserving, or even attacks on
democracy \cite{Wylie__Mindfuck__2019}. Secondly, artificial intelligence
programs reason in a simplistic way, but the real world is complex, especially
in the legal field which leaves a certain part to the human interpretation of
the law and characterization of the fact. A machine learning program has great
difficulty in dealing with the unexpected events that happen in the real world.
Intelligent system algorithms are black boxes that are impossible to
understand, they are unregulated and difficult to question in the case of the
presence of bias, and in some cases they amplify inequalities
\cite{ONeil__Weapons_of_Math_Destruction__2016}. Thirdly, when a classifier
learns about data  collected on past situations, it performs statistical
deductions and transform correlations between variables into implication
relationships. This can lead to problems with dramatic consequences such as
gender bias or racial discrimination \cite{Angwin_et_al__Machine_Bias__2016}.

We present in this paper a method to predict the spouses' alimony after a
divorce in France. To do this, we have a corpus of first instance court
decisions (formerly known in French as ``\textit{tribunal de grande
instance},''  now ``\textit{tribunal judiciaire}'') which have already been
subject to traditional analysis (manual data entry, statistical and econometric
analyses). This allows us, by working on the same corpus, to test the
reliability of the results obtained and to exceed the methodological limits
encountered. This paper is organized as follows: Section~\ref{sec:motivations}
provides information on the interest of the various stakeholders in advancing
the knowledge of the determinants of court decisions.
Section~\ref{sec:related-work} presents a state of the art, not only on the
technical progress made in the field of machine learning and law and predictive
judicial analytics, but also on the ethical difficulties identified. We present
our project on the study of the knowledge extracted from jurisdictional
productions in Section~\ref{sec:project}, how this study relates to the
analyses on the determinants of the economic consequences of a divorce, and how
this knowledge can be used for the design of decision support tools. An alimony
prediction model is proposed in Section~\ref{sec:model}, followed by the
results of its application to the available data in Section~\ref{sec:expe}. We
discuss the results obtained in Section~\ref{sec:discussion}.

\section{Motivations}  \label{sec:motivations}

The study we propose in this paper focuses on the prediction of the alimony
after a divorce. An \textit{alimony}, also known as \textit{spousal support},
is defined as the ``transfer of income between spouses intended mainly to
reduce inequality in living standards following a divorce''
\cite{Bourreau-Dubois2017}. Our objective however is not only to produce a
simple predictive model. This work is motivated by a differentiated
contribution that we can have for the different types of actors concerned by
the subject. The various stakeholders have indeed every interest in advancing
in the knowledge of the determinants of court decisions, but their motivations
differ according to their position.

First, for the litigants --whose positions are known through interviews with
their lawyers-- what is most important is to have an idea of what they can
expect from the court decision. Applied to the economic consequences of a
divorce, it is a question of whether they can expect to receive (or pay) an
alimony and how much it will be. This information not only makes it possible to
plan for the future but also to base more global negotiations on the whole of
the consequences of the divorce. However, by providing a list of
non-exhaustive, non-prioritized criteria, and sometimes by referring to facts
that are difficult to establish, the drafting of the law makes alimony one of
the elements of the divorce decision the most difficult to anticipate.

Second, for the lawyers, the motivation relates to the need to respond to the
predictability concerns of their clients, and thus to show that they have
mastered the subject. They must also establish a judicial strategy to defend in
the best interests of divergent interests, either by helping the divorcees to
reach an agreement, or by making a legal claim. In either case, it is important
to have objective criteria by which to provide guidance to their clients on
what they can expect from a court decision. However, they know that the hazard
is important too. They use therefore increasingly decision-making tools such as
scales or guidelines created by practitioners, predictive judicial analytics
tools \cite{Chen__2019} put on the market by private companies
\cite{DeJong__2019}, or databases which provide them with quantified case law.

Third, for the judges, there is no consensus: they are divided on the
advisability of using decision support tools, while recognizing that fixing
alimony is difficult. They are generally very attached to their appraisal and
they consider these tools to be optional, but they still use the
decision-support tools that are the scales created by practitioners
\cite{Sayn_et_al__2019}. The use of predictive judicial analytics tools, which
relate to the analysis of large amount of court decisions, is seen as
potentially affecting their freedom to decide, by allowing judge profiling
(prohibited by French law). However, the desire to produce comparable decisions
for comparable clients' cases is very present, especially for judges who assume
managerial responsibilities within the jurisdiction: they must pay great
attention to the importance of the regularity of the decisions rendered in
their jurisdiction. For jurisdictional organizations, predictability is also
considered as the means to favor agreements and unclog the courts.

Fourth, from the research point of view, the production of knowledge is an end
in itself. Knowing the determinants of court decisions is part of a realistic
approach to the law, in which judges have a prominent role in the application
of general and abstract rules to particular situations. It is a question of
better knowing the decision-making mechanisms and of identifying not only how
the legal criteria for decision are used but also to know if other criteria
interfere in the decision of justice, i.e., bias or extra-legal factors.
However, traditional analyses are very cumbersome to implement (manual entry)
and not always sufficiently efficient for the identification of the
determinants of decisions (statistical and econometric analyses). The use of AI
is considered here as a way to overcome this methodological bottleneck. For
researchers in computer science and mathematics, the objective is different but
convergent: to be able to develop new predictive models. In this project,
collaboration between disciplines is of course essential.

\section{Related Work} \label{sec:related-work}

Intelligent algorithms are applied in the legal field from the beginnings of
artificial intelligence with the use of expert systems in the late 1980s
\cite{Bench-Capon_et_al__2012}. When machine learning techniques have been
used, it was mainly the methods giving understandable models that have been
favored, such as rule-based approaches or dictionary-based models. Decision
trees \cite{Quinlan__1986} and random forests \cite{Breiman__2001}, as well as
techniques derived from them (e.g., extremely randomized trees
\cite{Geurts_et_al__2006}), have been widely used for predictive judicial
analytics  purposes
\cite{Katz_et_al__Predicting_Behavior_Supreme_Court_USA__2014}. More recently,
the use of Natural Language Processing techniques (such as N-gram features
obtained with a Bag-of-Words model) combined with statistical approaches (e.g.,
SVM \cite{Vapnik__SVM__1998}) have also shown very good results in predicting
court decisions \cite{Aletras_et_al__2016}. Deep Neural Networks have also been
applied in legal analytics in recent years, replacing more traditional
techniques that required expensive manual processing and only achieved poor
performance \cite{ONeill_et_al__2017}. The \textsc{word2vec} model
\cite{Mikolov_et_al__Word2vec__2013}, with the skip-gram and Continuous
Bag-of-Words (CBOW) algorithms, is capable of finding semantic similarities on
the basis of the co-occurrence of terms in large corpora of documents. By using
a legal corpora from various public legal sources for training such a model, it
is now possible to use a Law2Vec model to provide the semantics associated to
legal words in English \cite{Chalkidis_Kampas__2019}.

There are many machine learning techniques used in the law, but what do people
really want? In addition to greater efficiency in the legal process, which is a
stressful but also costly and time-consuming process, the answer to this
question depends on the type of stakeholder \cite{Muhlenbach_Sayn__2019}. The
use of machine learning is also motivated by the fact that a machine is
supposed to not be sensitive to the same extra-legal factors as a human being,
as can be judges who are more or less lenient in their decisions depending on
the time of day and what they ate
\cite{Danziger_et_al__Extraneous_factors_in_judicial_decisions__2011}. In
addition, judges are expected to apply the law in the same way, regardless of
their personal value scales, sensitivities, or political orientation
\cite{Cohen_Yang__2019}. Nevertheless, inter-judge disparities in predictions
are high, so much so that it was possible to predict the outcome of a trial
with a fairly good success score by taking into account as variable only the
surname of the judges that try the case \cite{Medvedeva_et_al__2020}. Thanks to
sentencing guidelines, it is fortunately possible to reduce the disparities
between judges \cite{Bourreau-Dubois2020}.

Human judge decisions are not pure: they can be biased, whether the judges are
aware or not of these biases and extra-legal factors. However, since machine
learning algorithms are based on court decisions that contain biases, it is to
be expected that the classification models they produce will also be tainted
with these same biases. Scattered within a few connection weights between
neurons lost in a deep neural network or associated with a variable that will
play a role in removing the model from the legal framework, such a bias can be
extremely difficult to find, which makes the source of the problem difficult to
identify, and preventing any rational and justified explanation in a court
\cite{Barocas__2016}. Different strategies have been studied to combat these
biases.  On the data side, to counter the problem of unbalanced data which
tends to reduce the chances of people from minorities in decision-making
problems (e.g., remission of sentences, access to consumer credit, selection of
an application for a position), studies suggest collecting more data for
increasing the sample sizes of these minorities \cite{Chen_et_al__2018}. On the
learning algorithm side, traditional methods have been adapted to deal with
these biases, such as modifying Naive Bayes classifier in order to perform
discrimination-aware classification \cite{Calders_Werwer__2010}. In addition,
work has been specifically devoted to neutralizing learning biases that pose
ethical problems, for example ``race neutral'' predictive modeling of decisions
on pre-trial release and paroling \cite{Lum_Johndrow__2016,Johndrow_Lum__2019}.
Finally, even for models known to be considered as black boxes such as deep
neural networks, work has been done to place around them a ``glass box'' by
mapping moral values into explicit verifiable norms that constrain the inputs
and outputs, so that these if they remain in the box, it is guaranteed that the
system adheres to the value \cite{Tubella_et_al__2019}.

We can say that, in general, there has been a clear increase in work in the
field that has addressed the societal repercussions that machine learning
models could have. Many works are no longer just focussing on the prediction
accuracy, but also on fairness and equality before the law, on transparency and
accountability, and on informational privacy and freedom of expression
\cite{Scantamburlo_et_al__2018}. It must be said that these problems have
generated strong reactions, both from civil societies and associations, but
also from nations. Regarding this concern for ethical issues related to the use
of artificial intelligence and machine learning, we can mention in particular
the drafting of the \textit{Asilomar AI
principles}\footnote{\url{https://futureoflife.org/ai-principles/}} in the USA,
the \textit{Montr\'{e}al Declaration for a responsible development of
Artificial Intelligence}\footnote{\url{https://tinyurl.com/y3ban2eq}} in
Canada,  the \textit{Villani Report ``For a meaningful Artificial
Intelligence''}\footnote{\url{https://www.aiforhumanity.fr/en/}} in France, or
the \textit{Ethical charter on the use of Artificial Intelligence in judicial
systems and their environment}\footnote{\url{https://tinyurl.com/y9tknlba}} in
the EU.

\section{Study of the use of knowledge based on jurisdictional productions for the design of decision support
tools} \label{sec:project}

The study relates to the alimony prediction in France. This specific focus
allows to understand the court decisions and see how they are conceived on this
particular question.

Following a partnership with the French Ministry of Justice, thousands of court
decisions dating from the year 2013 covering dozens of first instance courts
were collected and analyzed. At the time of the study, in France, there were
173~trial courts of this type (i.e., one or more per French department). These
court decisions have therefore already been the subject of a first analysis:
the researchers developed a data entry grid based on the reading of part of
them and then proceeded to the data entry from one-to-one reading of court
decisions. The database created in this way was finally subjected to
statistical and econometric analyzes. With this first study, the researchers
conclude, for example, that the duration of the marriage or the incomes of the
spouses are the determining factors. In the calculators found online to assess
the amount of an alimony --in particular for the different states of the United
States or the different provinces of Canada--, we find equivalent results: the
marriage length in years and the gross incomes of the two spouses are the data
always requested.

The study was not conducted with a purely predictive objective. The idea behind
this work was rather to report on the processes followed by the judges to make
their decisions, and more particularly concerning the following points:

\begin{itemize}

\item find the determinants allowing to indicate that a litigant (a former spouse) is eligible or not for the
alimony;

\item find the determinants used to calculate the amount of this alimony;

\item analyze the determinants in order to see if there are hidden extra-legal factors among them.

\end{itemize}

This study indeed makes it possible to identify if there are extra-legal
factors that must be integrated into the model to understand the way in which
judges make their decisions (e.g., if there is an effect of the judge, if there
is an effect of the lawyer, or even an effect of the court). As the first
instance courts are associated with a given jurisdiction, therefore with a
specific geographical area of France, it is interesting to see if there are
differences between the different seats of the courts.


\section{Alimony Prediction Model} \label{sec:model}

Consider a corpus of first instance divorce court decisions codified in a
database, representing the legal production carried out at national level in
France, we conduct a predictive analysis which aims to:

\begin{itemize}
\item Step 1: predict the alimony eligibility and acceptance by the court;
\item Step 2: predict the alimony amount set by the court;
\item Step 3: adjust the alimony amount from step 2 by the outcome from step 1.
\end{itemize}

For the learning phase, the model is trained with court decisions in a
supervised way by using two submodels: a classification model in step 1, and a
regression model in step 2. We predict the adjusted alimony amount by
considering both its acceptance probability and the amount:
\[\widehat{y}_{alimony}=\widehat{y}_{C} \times \widehat{y}_{R}\]
where $\widehat{y}_{alimony}$ is the adjusted alimony amount and
$\widehat{y}_{C}  $, $\widehat{y}_{R} $ are respectively the predicted variable
of the classification and the regression model.

Note that $ \widehat{y}_{C}  $ is recoded to 0 for absence of alimony and 1 for
acceptance of alimony. This configuration makes it possible to cross the
variables which relate to the alimony eligibility and those on the alimony
amount. Although regression hardly gives exactly \textit{zero} as the outcome,
observed alimony amount could be \textit{zero} while the divorcing spouses are
not eligible to the alimony for example. Those cases disturb the regression as
the linear relationship assumption is not satisfied. By splitting the alimony
prediction model into two independent submodels, we can solve this problem.

\section{Experiments} \label{sec:expe}

\subsection{Dataset}

We validate our model using a database collected previously as part of a
collaboration with the Ministry of Justice. The database, made up of 5,453
divorce decisions, contains 3,203 cases for which the question of granting an
alimony arose and in only 2,678 of them ultimately an alimony was approved by
the court. It is therefore possible to calculate a success rate and to answer,
by comparing the two types of cases, a predictive question of the court knowing
the divorcing spouses situation and their alimony request.

The proposals for the alimony amount made by the divorcing spouses, if
mentioned in the decisions --which is not always the case-- can be expressed
either in terms of monthly payment or in terms of capital
\cite{Belmokhtar_Mansuy__2016}. We only have 280 cases (8\%) with a court
decision on the form of monthly payment, and we decide to not include those
atypical cases in our model.

Moreover, the database consists of two very distinct situation, after deletion
of the unusable cases. The first situation concerns the 1,524 cases where the
parties have agreed, here the offer is equal to the demand and this amount is
approved in more than 99\% of the cases by the judge. This systematic approval
explains the almost perfect estimate of the amount of alimony set by the judge.
The second situation concerns the 1,257 cases where the parties did not agree
neither on the amount nor the principle of the alimony. It is thus only on this
small subsample that the question of the estimation of the alimony from the
court decision is really relevant.

\subsection{Feature Selection}

The prospect that the ``\textit{Loi pour une R\'{e}publique num\'{e}rique}''
(``Law for a Digital Republic,'' known as the \textit{Lemaire Law}, 2016) opens
in the field of law, namely the free access of all French court decisions
digitized in the near future, would facilitate the application of machine
learning models. However, the court decision annotation is an expensive and
time-consuming process. Therefore, searching for important determinants makes
the model explainable and reduces the cost of data labeling for new court
decisions. We perform a feature selection for each of our submodels.

\subsubsection{Classification}


To identify the determinants, we use the most common algorithm with tree-based
models: the Gini importance \cite{Breiman_et_al__CART__1984}. This criterium
counts the number of times a feature is used to split a node, weighted by the
number of observations in the node. The 15 most important variables presented
in Table~\ref{classif-var} lead to the following observations: (1) The
professional situation of the divorcing spouses is determinant as both the
activity status and the income are important. (2) The appearance of the
variable ``Seat of First Instance  Court'' is surprising and should not take
place because the law should be the same throughout the French territory.

\begin{table}[t]
\caption{List of the most important features in classification whether to grant
alimony or not using Gini importance} \label{classif-var} \vskip 0.15in
\begin{center}
\begin{small}
\begin{sc}
\begin{tabular}{lr}
\toprule
Variables & Gini \\
\midrule
Activity status of the wife& 19.9\\
Activity status the husband &15.6\\
Salary of the husband&  30.5\\
Retirement pensions of the husband &13.4\\
Salary of the wife  &26.1\\
Other income of the wife    &10.6\\
Nb of children from the couple  &16.2\\
Nb of adult children of the couple  &13.7\\
Common life during marriage &21.4\\
Temporary support payments  &25.0\\
Temporary allocation of domicile &10.7\\
Capital paid at once requested &33.7\\
Type of capital in cash requested & 16.2 \\
Type of capital in cash offered & 18.5 \\
Seat of First Instance  Court & 107.1\\
\bottomrule
\end{tabular}
\end{sc}
\end{small}
\end{center}
\vskip -0.1in
\end{table}

The goal of our work is to build an ethical, unbiased model, without
extra-legal factors. Despite its importance, we decide to not use the variable
``Seat of First Instance Court.'' In order to overcome this problem, we use a
Random Forest classification with only the 14 variables left. After tuning our
predicting model, we obtain an accuracy rate of 99.89\% and an AUC of 0.999.
This almost perfect rate shows that using extra-legal factors like ``Seat of
First Instance Court'' is unnecessary.


\subsubsection{Regression}

We use a stepwise forward selection to identify significant features. We report
a multiple R-squared of 0.6619 and an adjusted R-squared of 0.6579 with an
Ordinary Least Squares regression \cite{Goldberger__1964}. Results in
Table~\ref{reg-var} lead to following observations: (1) The proposals for the
alimony amount made by the divorcing spouses (offered and requested) are
decisive as far as the judge must decide \textit{infra petita}. We can then see
that supply and demand almost perfectly explain the amount of alimony fixed. We
could therefore conclude that it is almost enough to know the proposals to
determine the amount withheld by the court. (2) Interim measures are temporary
measures taken by the judge to officer the conjugal and family life of the
divorcing spouses during the process of divorce. They are set at the time of
the conciliation hearing.\footnote{The programming and justice reform law
2018-2022 of March 23, 2019 modified the divorce procedure and notably
abolished the previously compulsory conciliation hearing for contentious
divorce (art. 22). This development, which will take effect on September 1,
2020, does not exclude the possibility of fixing provisional measures during a
first hearing.} They take effect from this date and end at the time of divorce.
These interim measures are good predictors of the alimony amount, in particular
the temporary pension offered during the non-conciliation order. Indeed, it is
common knowledge that some judges fix an alimony amount equal to a multiple of
the amount of temporary pension.

\begin{table}[ht]
\caption{List of the most important features in regression using forward
stepwise} \label{reg-var} \vskip 0.15in
\begin{center}
\begin{small}
\begin{sc}
\begin{tabular}{lr}
\toprule
Variables &  Estimate \\
\midrule
Intercept&  8403.15\\
Capital at once requested&  0.33\\
Capital at once offered&    0.79\\
Capital at once in a joint request& 0.88\\
Capital over time offered&  0.48\\
Capital over time in a joint request&   0.80\\
Capital over time requested&    -0.56\\
Months of capital over time requested&  353.42\\
Pension offered &   -88.72\\
Pension requested & 40.80\\
Temporary pension offered&  1.37\\
Salary of the wife& -7.86\\

\bottomrule
\end{tabular}
\end{sc}
\end{small}
\end{center}
\vskip -0.1in
\end{table}

From the Table~\ref{reg-var}, we can easily calculate the amount of
$\widehat{y}_{R}$ from the following equation:

$$ \widehat{y}_{R} =  \begin{array}{rrl}

 & 0.33 ~ \times & \textrm{Capital at once requested} \\
+ & 0.79 ~ \times & \textrm{Capital at once offered} \\
+ & \ldots  &\\
+ & 40.80 ~ \times  & \textrm{Pension requested} \\
+ & 1.37 ~ \times & \textrm{Temporary pension offered} \\
- & 7.86 ~ \times & \textrm{Monthly salary of the wife} \\
+ & 8403.15 & \\
\end{array} $$

With an intercept whose value is very different from \textit{zero}, it is very
difficult to get an null estimate of $ \widehat{y}_{R} $ (corresponding to a
non-allocation of an alimony by the judge). This motivates the interest of
having made a combination model between a classification method (giving a value
of 1 or 0) and a regression method.

\begin{figure}[ht]
\vskip 0.2in
\begin{center}
\centerline{\includegraphics[width=\columnwidth]{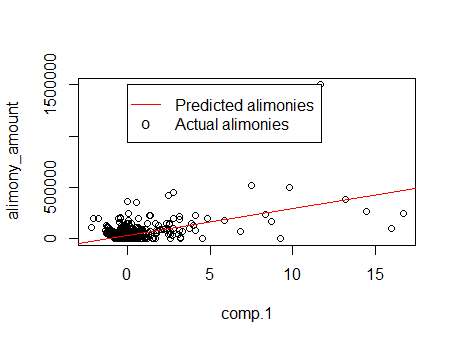}} \caption{Multiple
linear regression prediction of the alimony as a function of the set of
variables selected for the model (1st component of the PCA).}
\label{fig:prediction}
\end{center}
\vskip -0.2in
\end{figure}

The Figure~\ref{fig:prediction} presents the predicted value for the alimony
$\widehat{y}_{R}$ as a function of the set of the variables selected for the
model (Table~\ref{reg-var}). The figure is however not very representative of
the quality of the model: the first component of the principal component
analysis made on the dataset explains only 20\% of the variance in the data
(all the 11 continuous variables kept for the regression model are necessary
for the alimony amount prediction, but these variables are not correlated with
each other).

\subsection{Performance Comparison}

Our model performs best by fusing a Random Forest (RF) classification
algorithms with an Ordinary least squares (OLS) regression or a Quantile
regression. OLS regression models the effect of explanatory variables on the
mean value of predicted variable. As the mean is more affected by outliers and
other extreme data present in our database, we also use the quantile regression
\cite{Koenker__2005} which estimates the conditional median of the predicted
variable.

In order to measure the relevance of our predictive model, we have calculated
the absolute value of the difference between the actual value and predicted
value. We report the results distributions, called absolute errors, for each of
the two regressions by themselves and by application of our model in Table
\ref{performance}.

\begin{table}[ht]
\caption{Absolute errors in prediction (in thousands of euros)}
\label{performance} \vskip 0.15in
\begin{center}
\begin{small}
\begin{sc}
\begin{tabular}{lrrrr}
\toprule
Model &  Mean & Median & $ \sigma $ ~~ & $ R^{2} $ \\
\midrule
OLS Reg. & 21.46 & 10.64 & 35.93 & 0.66 \\
Quantile Reg. & 19.73 & 9.04 & 40.93& 0.62\\
RF$\times  $OLS Reg. & 16.46 & 3.95 &35.59 & 0.70\\
RF$\times  $Quantile Reg.& 15.95 & 3.43 &32.01& 0.65\\

\bottomrule
\end{tabular}
\end{sc}
\end{small}
\end{center}
\vskip -0.1in
\end{table}

The prediction absolute errors presented in Table~\ref{performance} lead to the
following observations: (1) Quantile regression gives less errors. (2) Our
alimony prediction model outperforms each regression. (3) A lower R-squared is
not inherently bad.

\section{Analysis and Discussion} \label{sec:discussion}

Another possibility for selecting the variables is based on the statement of
the Civil Code. The rules relating to the alimony are provided on articles 270
to 281 of the Civil Code. These provisions mention in particular its
calculation criteria, its terms of payment or revision of its amount, or even
the rules applicable in certain specific situations such as the death of the
debtor. We can cite the two most important.

Article 270: ``One of the spouses may be required to pay the other a benefit
intended to compensate, as far as possible, for the disparity that the
breakdown of marriage creates in the respective living conditions. (...)
However, the judge may refuse to grant such a service if equity requires it,
either in consideration of the criteria provided for in article 271, or when
the divorce is pronounced at the exclusive wrongs of the spouse who requests
the benefit of this service, in view of the specific circumstances of the
breakdown.''

Article 271: ``The alimony is fixed according to the needs of the spouse to
whom it is paid and the resources of the other, taking into account the
situation at the time of the divorce and its development in the future
predictable. To this end, the judge takes into consideration in particular:

\begin{itemize}
\item the duration of the marriage;
\item the age and state of health of the spouses;
\item their professional qualification and situation;
\item the consequences of the professional choices made by one of the spouses during the common life for the education of the children and the time that it will still be necessary to devote to it or to favor the career of his spouse to the detriment of his own;
\item the estimated or foreseeable patrimony of the spouses, both in capital and in income, after the liquidation of the matrimonial regime;
\item their existing and foreseeable rights;
\item their respective retirement pensions situation, having estimated, as far as possible, the reduction in pension rights that may have been caused, for the spouse claiming the compensatory allowance, by the circumstances referred to in the sixth
paragraph.''
\end{itemize}

Clearly, our feature selection technique do not give all of those legals
determinants the same importance especially when the reconstruction of these
indicators requires mobilizing dozens of information provided in the decisions.
It is not always easy to measure for example:\textit{``the disparity that the
breakdown of marriage creates in the respective living conditions.''} We
prioritize the statistical significance in our model in order to provide an
accurate prediction. Therefore, among those legal determinants, our
classification  managed to discover the duration of the marriage, the spouses'
professional situation, their incomes, and the husband's retirement pensions.
Meanwhile, our regression only takes into account the salary of the wife but
depends a lot on the supply and demand for alimony from both parties. It is not
surprising that the amount fixed by the judge reflects the requests of the
parties, the judge having the general obligation to rule in this context. The
results obtained show that the magistrates respect this procedural rule which,
in the end, prevails over the legal criteria of decisions. We can therefore
suggest that it is upstream, at the stage of developing requests, that these
legal criteria can play. Moreover, except the salary of the wife, variables
used for classification are not reused for regression in our model.


In term of performance, a median absolute error of only \euro{3,432.98} and a
variability of the alimony explained up to 70$  \%$ clearly show the superior
quality of our model compared to a regression. However, our model suffers from
a slight underestimation bias because the predicted alimony mean are
respectively \euro{28,826.45} for the OLS regression and \euro{24,142.24} for
the Quantile regression compared to the average alimony in our database of
\euro{33,653.89}. We note that the difference between the average alimony in
our database of \euro{33,653.89} and its median of \euro{15,000.00} is strongly
determined by extreme deviations.

Despite the globally satisfying quality of prediction, we cannot fully trust a
tool which, on average, offers a prediction generating an error that exceeds
half of the actual alimony amount.

\section{Conclusion and Further Work} \label{sec:conclusion}

The recent increase in the efficiency of machine learning algorithms has
allowed the arrival of new tools based on artificial intelligence. Since then,
new companies exploiting these tools and technologies have appeared all over
the world on the legal market space. Even if it seems to be emerging that AI
will not replace lawyers, it is likely that lawyers using AI-based tools will
replace traditional lawyers.

The work presented in this paper does not seek to produce an AI system capable
of making decisions automatically, possibly replacing lawyers. From our
perspective, AI is not used to make decisions but to provide information on
only part of what constitutes a court decision, relating to the setting of an
amount. In this context, the objective is not to design a machine capable of
following a reasoning allowing to reach an overall result (a decision) but only
to know those of the criteria which determine the amounts retained by the
magistrates, in the exercise of their discretion. This allows us to know and
understand the ways in which judges use the margin of freedom left to them by
the necessary incompleteness of the law. It is above all a question of
knowledge, allowing both to show the judicial uncertainty, to explain it and to
detect if there is any implicit bias in action. The distinction in the analysis
between the legal determinants of the decision and the non-legal determinants
pursues this objective. This knowledge can also be a lever for action, by
making it possible to offer professionals a decision support tool. The
assumption is that such tools would have an effect on practices, as long as
they were fairly widely used. However, it is not a question of freezing these
practices, not only because decision-making tools can remain optional but also
because their mastery by professionals leads to their development. It is
therefore not a question of blocking the future from data taken in past
decisions, even when the legal and social context is changing, but of giving
ourselves the means to steer desirable developments.

In the specific area of the alimony studied here, such an evolution may be
desirable, for several reasons. On the one hand, at the stage of the judicial
decision (and without prejudging what happens at the stage of the preparation
of the requests), it has been shown that the legal criteria supposed to
condition the amounts withheld only intervene on the principle of allocation of
a service. On the other hand, the logic behind the work before the referral to
the judge remains poorly understood, as shown by the tools already used by
practitioners, both numerous and very different from each other. Questions
remain unanswered: what are we trying to compensate for? are legally legitimate
claims for benefits still being made? how do the parties and their lawyers
determine the requests they make? It is therefore not necessarily appropriate
to deduce from the only data presented here an operational scale, even if the
data were sufficient. Moreover, this data can be invaluable in constructing a
scale which effectively helps the magistrates and the parties to fix comparable
amounts in comparable situations.

In the continuation of our work, we plan to develop systems for automatic
analysis of raw texts of court decisions in order to directly detect the values
of interest in the text (with NLP and text mining techniques), avoiding the
long and tedious phase of manual document analysis. A useful model must
represent suitably the behavior of what it is supposed to model. It is
therefore necessary to carry out a regular update by feeding the model with new
examples of court decisions, reflecting the decision-making mechanisms of
judges and the way they have to make their decisions according to the evolution
of society and changes in the law. For example, same-sex marriage in France has
been legal since 18 May 2013. So there are now also possibilities for same-sex
divorce, thus leading to cases of determination of alimony hitherto not
encountered.

\bibliography{ICML2020-LML-MNPS}
\bibliographystyle{icml2020}

\end{document}